\documentclass[twocolumn]{article}

\usepackage[utf8]{inputenc}

\usepackage{comment}

\usepackage{graphicx}

\graphicspath{{./figures/}}

\title{Problem examination for AI methods in product design}

\author{Philipp Rosenthal, Oliver Niggemann}

\date{August 19, 2021}

\begin{document}

\maketitle

\begin{abstract}

Artificial Intelligence (AI) has significant potential for product design: AI can check technical and non-technical constraints on products, it can support a quick design of new product variants and  new AI methods may also support creativity. But currently product design and AI are separate communities fostering different terms and theories. This makes a mapping of AI approaches to product design needs difficult and prevents new solutions.

As a solution, this paper first clarifies important terms and concepts for the interdisciplinary domain of AI methods in product design. A key contribution of this paper is a new classification of design problems using the four characteristics decomposability, interdependencies, innovation and creativity. Definitions of these concepts are given where they are lacking. Early mappings of these concepts to AI solutions are sketched and verified using design examples. The importance of creativity in product design and a corresponding gap in AI is pointed out for future research. 
    
\end{abstract}

\section{Motivation}

Driven by advances in computer technology, Artificial Intelligence (AI) is becoming more important in recent years~\cite{BundesministeriumfurBildungundForschung.2020}.
This is also true for the domain of product design and development. The design of a product determines how good a product can be manufactured, marketed and maintained throughout its life cycle. With shorter development cycles driven by globalization efficient designs and design processes become more important 

AI can be helpful in several ways to support the design engineer. I can support by finding a vast number of solutions to a design problem as well as support in repetitive tasks like routine design. However, AI is not yet able to replace experience and creativity of human engineers. There is for example no general algorithm which solves the central problem of functional decomposition for any given design problem. This is in part due to the subjective nature of functional decomposition. Creativity is an important aspect for solving new design problems. So far all creativity in AI methods for solving design problems can be seen as induced by the programmer.

\smallskip
To identify potential AI applications in product design, the following research questions (RQ) must be answered:

\noindent \emph{RQ 1: Terms} Which terms describe problem and solution approaches in design and AI and can they be harmonized?

\noindent \emph{RQ 2: Design/AI Mapping} Can we define characteristics of product design problems, so that these characteristics correspond to the applicability of specific AI algorithms?

\noindent \emph{RQ 3: Evaluation} Can we verify the new characteristics from RQ 2 by analyzing existing design examples? 

The sections \ref{terms} and \ref{class} address the fundamental terms in product design relevant to AI and the main directions AI has taken in in the design domain. Section \ref{characteristics} identifies problem characteristics. The following section takes examples from the the AI directions in product design and showcases the identified characteristics. The final section gives a summary, answers the research questions and gives an outlook for future research.

\section{Important terms in product design}\label{terms}

This section clarifies fundamental terms of product design and puts them into perspective for AI. 

\smallskip
\noindent\emph{Product Design:} Product Design (Product Development) is the purposeful evaluation and application of research results and experience. The aim is to obtain technical products, software programs, materials, principle solutions and similar concepts.~\cite{VereinDeutscherIngenieure.1993}

\smallskip
\noindent\emph{Design Synthesis:} In regard to the design domain, the term synthesis can be seen from different points of view. Roozenburg~\cite{Roozenburg.2002} gives a good overview over the different senses of the term. In general synthesis is understood as "putting something together" or as "collection". Analysis as the opposite of synthesis can be expressed as the resolution of objects into their elements. Synthesis is also a phase of the design process in which possible solutions for subproblems are found to build a complete design from. Here Analysis is the phase of identifying the subproblems. The VDI 2221~\cite{VereinDeutscherIngenieure.1993} interpretes synthesis as a compilation of a list of designs which have to be evaluated to find the most suitable one for the given design problem. Synthesis as the assembly of subsystems by combining parts to a new whole and the functional analysis represented by e.g.~\cite{Pahl.1999} is also a prominent view on the term "synthesis".~\cite{Roozenburg.2002}

Gagan et al.~\cite{Cagan.2005} give a more computer centered description of synthesis in design. They state that design synthesis is the algorithmic creation of designs. A computer helps creating designs by organized, methodological modeling, implementation and execution.

\smallskip
\noindent\emph{Function and Behavior:} The most important aspect of a product is its function. Chakrabarti~\cite{Chakrabarti.1998} defines function as intentional behavior. The intention is in this meaning distinct for being something actual or expected. E.g. wood swimming on water is in that regard a behavior, transporting goods on water is a function since it implies intention. Deng~\cite{DENG.2002} points out that function can be implemented with behavior. To continue the example, a raft can be build out of wood to transport goods.

Deng~\cite{DENG.2002} also makes an important distinction between purpose functions and action functions. A purpose function is an abstract and subjective description of intention of a product a designer has. He defines action functions as abstractions of intended and useful behavior. Purpose functions can be implemented via actions functions.

The third important concept of Deng~\cite{DENG.2002} is the distinction in function representation. Functions can be either semantically of syntactically represented. The latter one makes functions accessible to computers and is thus of interest for AI.

\section{Classification of computational design synthesis}\label{class}

Chakrabarti et al.~\cite{Chakrabarti.2011} point out three major directions AI has taken in product design. Which are functional-based synthesis, grammar-based synthesis and analogy-based design. All three are addressed in the following.

\smallskip
\noindent\emph{Functional-Based Synthesis:} The process of this approach is functional decomposition in conjunction with finding partial solutions to these subfunctions. The subsequent synthesis is the most crucial part since it defines the total function the product needs to fulfill in order to satisfy the desired requirements.

Chakrabarti and Bligh~\cite{Chakrabarti.1994,Chakrabarti.1996,Chakrabarti.1996b} propose an approach to mechanical conceptual design using functional synthesis.

\smallskip
\noindent\emph{Grammar-Based Synthesis:} Grammatical Design consists of three parts. First there needs to be a vocabulary of elements. The second part is a set for transformation rules and the final needed part is an initial structure to which the transformation rules can be applied~\cite{Brown.1997}. The just mentioned grammar rules can be described as LHS $\rightarrow$RHS. Where the left hand side (LHS) resembles the location where the rule is applied in the design and the right hand side (RHS) defines the design transformation.~\cite{Konigseder.2014} 

Grammatical design is also known as design language. Design grammars are able to generate a wide variety of alternative designs to a design (sub-)problem. This generation is an important aspect of classical conceptual product design where the best evaluated solution is picked from a variety of design proposals. 

According to Chakrabarti et al.~\cite{Chakrabarti.2011} the two most prevalent grammars in design are graph and spatial grammars. There are six main steps which should be followed when engineering a design grammar. At first the representation has to be set, then the vocabulary needs to be defined. The third step is defining the grammar rules, followed by defining the initial design. The fifth step is the actual generation of design, followed by the final step of modifying the design grammar.

\smallskip
\noindent\emph{Analogy Design:} Gero~\cite{Gero.2000} classifies analogy (among with combination, transformation, emergence and first principles) as a process for creative designing. He views analogy as a product of processes in which specific aspects of one problem are matched and transferred to another problem.

This is inline with Chakrabarti et al.~\cite{Chakrabarti.2011} who differentiate analogy based design further into case based design (CBD) and biology inspired design (BID). CBD is a type of case-based reasoning (CBR). Here, the case which is most similar to the new target problem is retrieved from a case-base. This case then helps to solve the new problem.~\cite{Kolodner.1992} The similarity is crucial to CBR and can be expressed by a similarity function which maps the distance between the target problem and the cases in the case-base to a real number. The retrieved case is then reused to solve the given problem. In addition the solution is then revised to better fullfill any special requirements given by the problem. Ideally the new solution is then retained as a case in the case-base for future problems.~\cite{Hullermeier.2007}

BID can be used in the same way where the case-base consists of biological solutions to design problems. Databases containing biology inspired solution may also support functional-based synthesis.

An example for CBD would be the design of a truck by using analogical knowledge about an already designed car. BID has a prominent 
example in the development of airplanes. The necessary lift of objects was inspired by birds and how they create lift to fly.

\section{Characteristics of Design Problems}\label{characteristics}

A design problem is mainly characterized by the product-to-be-designed. The following subsections use product characteristics to characterize design problems. The main goal is to identify an alignment between these characteristics and suitable AI algorithms, this will be described in section \ref{usecases} using exemplary use cases (RQ 2).

\subsection{Characteristic 1: Decomposability}

A popular classification for design problems is complexity. One may say that designing a car is more complex than designing a bicycle for instance. This may be an obvious observation when comparing problems from the same or similar class of design, in the above case personal transportation. However, it becomes vague when comparing problems from different classes of design. E.g. is a bicycle more complex than a bridge? A rational start is examining the function structure of a given design problem. In order to do that the overall function of the desired product must be accessible for functional decomposition. 

According to Roozenburg~\cite{Roozenburg.2002} this is not always the case. When considering spatial relationships, systems in which the parts form an organic whole, any change in a subsystem providing a subfunction may impair the overall function. He calls these systems associate systems in contrast to flow systems, which are decomposable. 

If a design problem and with it its function is very trivial, decomposition might also not be possible, simply because there exists no subfunction to the overall function. If a device has to be designed whose sole purpose is to absorb tensions, then there is no meaningful subfunction to it.

This leads to the binary dimension of decomposability. A design problem is either decomposable or it is not. Examples for the former are the aforementioned transportation devices like bicycles and cars. A representative of the latter is a bridge or rope.

\subsection{Characteristic 2: Function Interdependencies}

\begin{figure} [h]
    \includegraphics [width=0.5\textwidth]{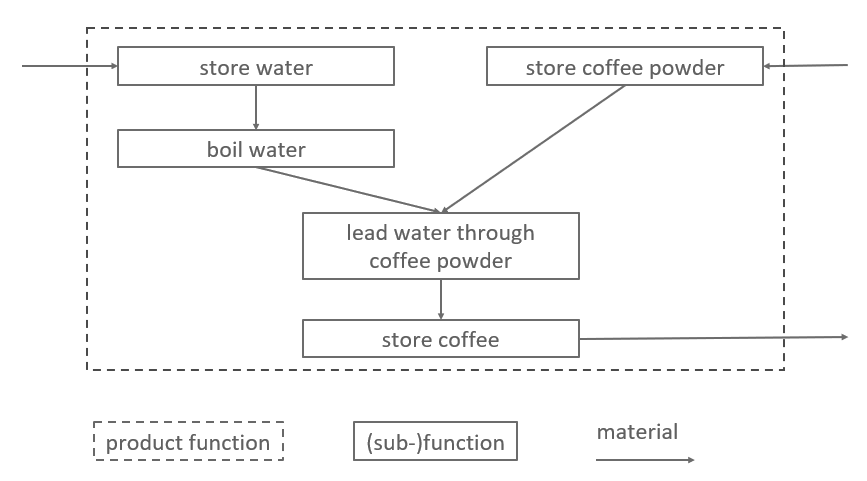}
    \caption{function structure - Here the overall function of \emph{make coffee} is decomposed into several subfunctions. The subfunction \emph{lead water through coffee powder} has an in-degree of 2 (boiled water and coffee powder) and an out-degree of 1 (coffee)}
    \label{fig:fktstr}
\end{figure}

If a design problem falls into the decomposability category, a function structure can be generated. This structure is usually represented by a directed acyclic graph. An example for such a graph is shown in figure \ref{fig:fktstr}. 

A large number of vertices or edges alone is not a good measurement for problem complexity, instead complexity is mainly causes by interdependencies between functions. And furthermore, a mere series of linear interdependencies, i.e. one predecessors and one successor, has a much smaller complexity than functions with high in-degrees or out-degrees. As a result the (nonlinear) interdependencies of a problem can be given as:

\begin{equation}\label{pc}
    PI=\frac{number\ of\ vertices\ with\ a\ degree\ >\ 2}{number\ of\ all\ vertices}
\end{equation}

\begin{figure} [h]
    \includegraphics [width=0.5\textwidth]{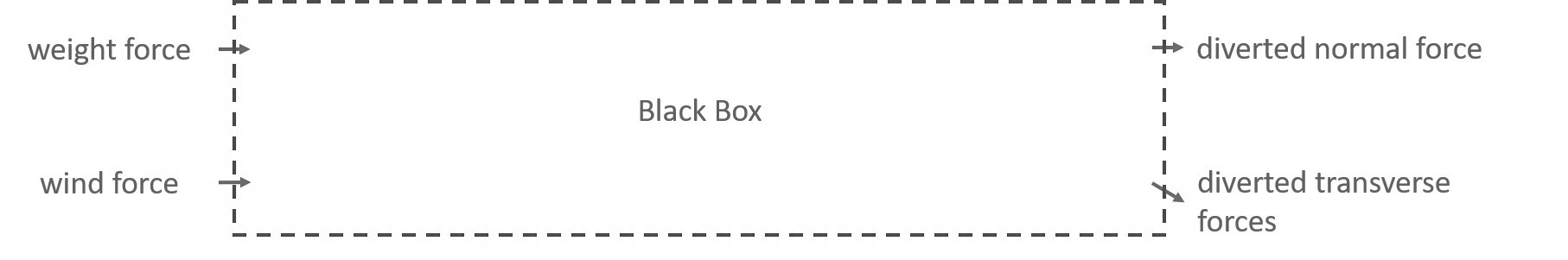}
        \caption{Bridge as a Black Box - Here the overall function of \emph{bear (dynamic) force} is not decomposed. The function has an in-degree of 2 (weight force and wind force) and an out-degree of 2 (diverted normal force and diverted transverse forces)}
    \label{fig:bbb}
\end{figure}

Where \(PI\) denotes the (nonlinear) design problem interdependencies. This also leads to two conclusions about nondecomposable problems. First, if such a problem has multiple inputs and/or outputs, its \(PI\) equals one. This is because the black box representing the overall function is a single vertex. Figure \ref{fig:bbb} gives an example for such a black box. 

Second, if a nondecompasable problem exhibits only a single input and a single output, its \(PI\) will be zero. Such problems usually represent fundamental functions like the rope mentioned above.

\subsection{Characteristic 3: Novelty of Design Problems}

Intuitively, re-designing a product is easier than inventing something new. To formalize the term ''new'', Gero~\cite{Gero.2000} divided the design process into several mutual exclusive categories. These types can be used to describe how "new" a design problem is. A first distinction can be made between routine and nonroutine design. Routine design includes all design problems for which all necessary knowledge needed to find a solution is known a priori. Nonroutine design is distinct to that. Here not all knowledge needed to find a solution for the design problem is known right away. In addition, nonroutine design can be further devided into innovative design and creative design.~\cite{Gero.2000} 

To formalize these fuzzy notions, some definitions are needed:

\noindent \emph{Variables}: Design knowledge is modeled based on variables. Each variable describes an, often physical, property of the product to-be-designed. Variables are often grouped into  subfunctions. E.g. when a subfunction correspond to lifting an object, the corresponding variable may describe properties of a rotor such as size and amount of rotors.

\noindent \emph{Design Knowledge}: Design knowledge are constraints on the variables, e.g. given in form of (in)equalities or in form of rules. 

According to Gero~\cite{Gero.2000} Innovation in design occurs when unexpected values for variables in design become possible.  An example for innovative design is to deduce a quadrocopter from a helicopter, the latter has an amount of one device that fulfills the function of applying lift to the object - a rotor in this case. While also having one device to counteract the resulting torque to the rotor axis. A quadrocopter is innovative because it extends the number of lifting devices beyond the prior known constraint and also reduces the number of devices needed to apply horizontal force. Analogical to \(PI\) a measurement for innovation can be given by:

\begin{equation}\label{i}
    I=\frac{number\ of\ variables\ with\ unexpected\ values}{number\ of\ all\ variables}
\end{equation}

Creative design is even a step beyond innovation. Here the variables  are (partially) unknown a priori and new variables are introduced into the design~\cite{Gero.2000}. Before Guglielmo Marconi's work on radio transmission, signals could only be transmitted by wire or visually. The introduction of radio waves can be seen as a creative solution to the problem of information transmission. Here the subfunction comprises a new variable "radio wave". As with innovation, a computational measurement for creativity can be given by:

\begin{equation}\label{c}
    C=\frac{number\ of\ new\ variables}{number\ of\ all\ variables}
\end{equation}

Boden~\cite{Boden.2009} points out a few more important aspects of creativity which should also be applied to innovation. An important constraint to creativity is that creative solutions have to be valuable. Not all alterations of values or variables result in a feasible solution. If that were the case, innovation and creativity would be trivial concepts which could be implemented by a randomizer. The tricky part is judging which alterations are imbued with value. 

The other aspect Boden~\cite{Boden.2009} mentions is the range of creativity. It can be distinguished between psychological creativity (p-creativity) and historical creativity (h-creativity). The former applies when the solution is new to the individual that created it and the latter applies when in addition the solution is new to the world.

Once unknown values or new variables are introduced into a class of design problems, these become known knowledge. In conclusion innovative and creative design can only be examined in retrospective where the innovation and / or creativity was unknown.

\section{Use Cases and Mapping to AI Algorithms} \label{usecases}

This section takes examples representing the aforementioned AI directions in product design from research and incorporates them into the described problem classification.

\subsection{Grammar for Shaft Design}

To demonstrate the mode of operation for design grammars Brown~\cite{Brown.1997} introduced the problem of designing a stepped grooved shaft. the overall function of such shafts is to forward torque while also diverting axial and radial forces. From this description a decomposability into basic subfunctions lies at hand. 

\smallskip
\noindent \emph{Classification into Categories:}  
The example of a shaft is not complex, since subfunctions mostly have only one input and one output, e.g. torque. The resulting \(PI\) equals zero. Even though the sections may also serve for e.g. bearing forces, there is no complexity since the shaft itself does not create interdependencies. For the vocabulary for this particular example, Brown states two shapes, being grooved and ungrooved and two labels and five simple transformation rules. Innovation occurs, when variables, in this case, section length, diameter and number of sections have unexpected values. 

It is possible, that the given grammar produces designs for shafts with unknown values for variables but their usefulness and with that their innovativeness can be argued. Since for design grammars themselves the vocabulary needs to be known a priori, creative solutions are impossible, since a design grammar is not able to introduce entirely new variables to a design, e.g. a square diameter section for a shaft which has no bearings.
\begin{figure} [h]
    \centering
    \includegraphics [width=0.25\textwidth]{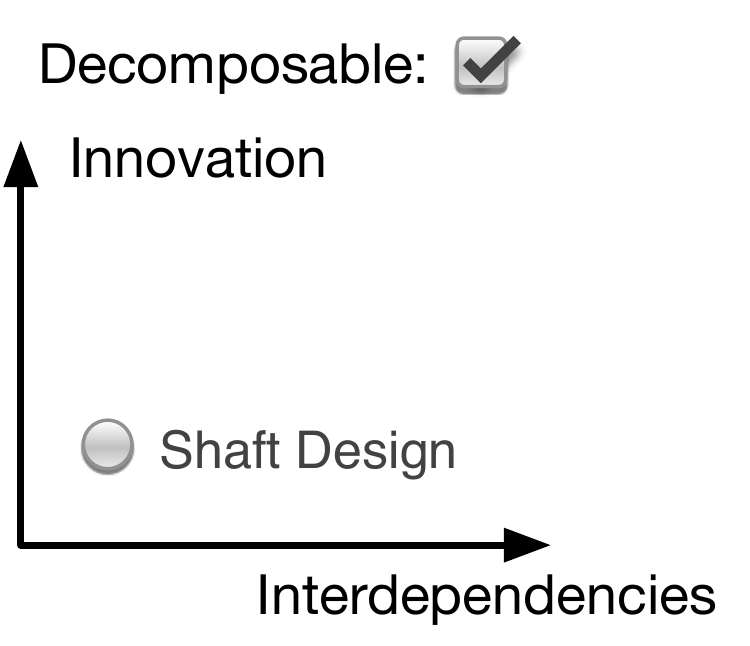}
    \caption{Categorization of the Shaft Design Example}
    \label{cat1.pdf}
\end{figure}

\smallskip
\noindent \emph{Potential AI Approaches:} 
In AI, such  knowledge models can be captured e.g. using rules~\cite{EASTMAN20091011}. Nowadays, often machine learning is used to either extract those rules~\cite{KO2021101620} from data. 

\subsection{Grammar for Gearboxes}

Holder et al.~\cite{Holder.2019} introduce a design language to synthesise gearboxes including 3D component arrangement. The overall function of a gearbox is to transmit torque. Subfunctions needed to synthesise this function are e.g. bearing of shafts, transmissions or reversal of torque direction. 

\smallskip
\noindent \emph{Classification into Categories:} 
As a result the problem of designing a gearbox is decomposable due to its identifiable and separable subfunctions. Responsible for the \(PI\) of gearboxes are functions which require to have radial or axial forces along with torque as input and corresponding outputs like diverting forces. Such a function may be represented by a bearing. When also considering 3D placement of parts the \(PI\) grow due to constraints which may add subfunctions to the design problem. E.g. a very narrow installation place may lead to additional gearwheels to achieve the required transmission. As with the shaft example, innovation may be achieved by an unexpected value for the given variables, e.g. number of gearwheels or the transmission between gearwheels. Depending on the given vocabulary of the design grammar an additional change in torque direction is also thinkable. Also creativity is not possible to achieve since all possible variables have to be included into the knowledge base being the vocabulary of the design language.
\begin{figure} [h]
    \centering
    \includegraphics [width=0.25\textwidth]{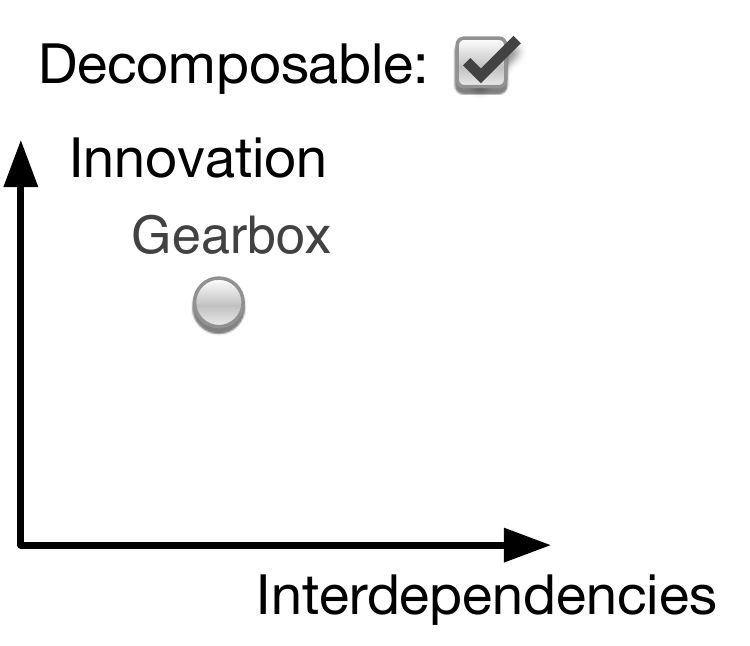}
    \caption{Categorization of the Gearbox Example}
    \label{cat2.pdf}
\end{figure}

\smallskip
\noindent \emph{Potential AI Approaches:}  
Such grammars~\cite{10.1115/96-DETC/DTM-1511} are another AI approach which is an extension of rule based approaches. By defining an order of symbols, geometric and other dependencies between subsystems can be modeled. Furthermore, their generative nature fits well to design~\cite{Konigseder.2014}. In the last years, learning of grammars has become more and more popular~\cite{Gero1994EvolutionaryLO}. 

\subsection{Analogy Based Coil Winder Design}

McAdams and Wood~\cite{McAdams.2002} approached analogical design using a similarity metric. They chose a coil winder for guitars as a problem to test the analogical design. This problem is well suited to demonstrate the characteristics of design problems which can be solved using the analogy approach. 

\begin{figure} [h]
    \includegraphics [width=0.5\textwidth]{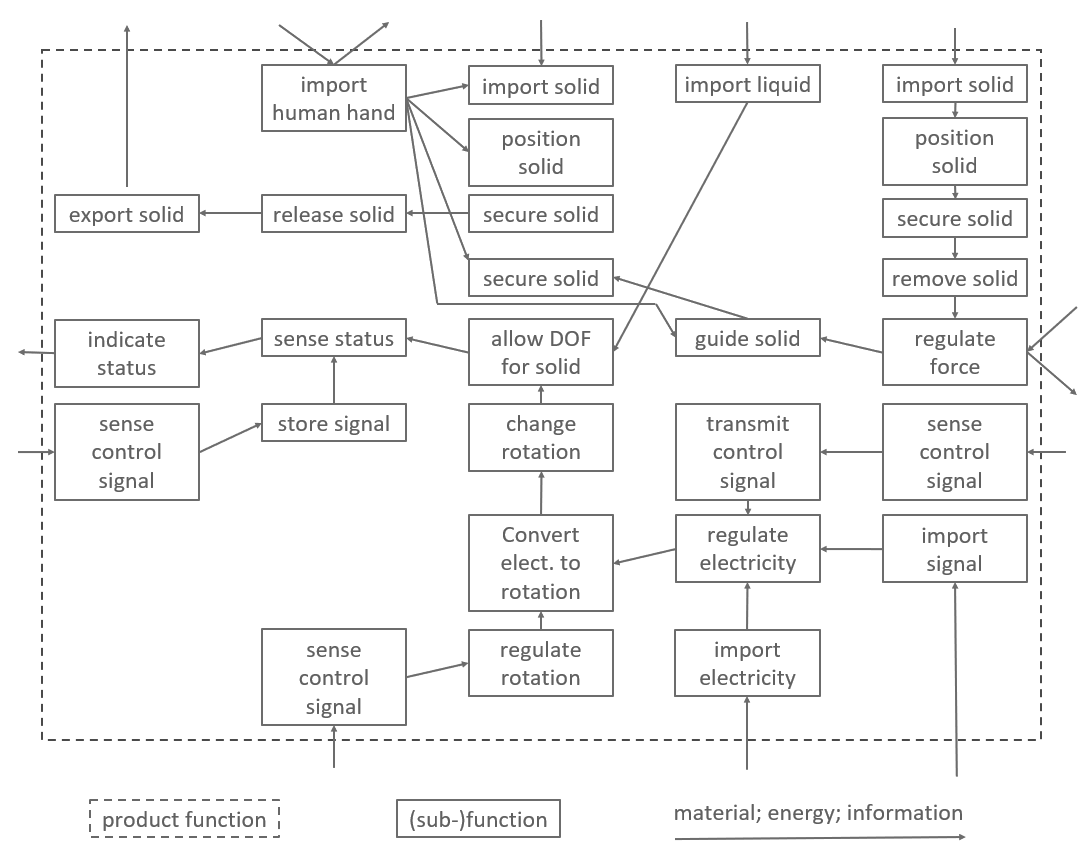}
    \caption{simplified Coil Winder function structure based on~\cite{McAdams.2002}}
    \label{fig:cw}
\end{figure}

\smallskip
\noindent \emph{Classification into Categories:} 
First the design problem is decompasable. Figure \ref{fig:cw} depicts  the function structure of such a device. From the function structure the \(PI\) measurement can be derived. In the shown example 12 of 28 nodes have a degree greater than two, in conclusion the \(PI\) has a value of \(3/7\).

Analogy based design solutions are open to creativity, as long as the analogies be it from the problem domain or not are included in the knowledge base. This implementation is usually done by the programmer. In conclusion the creativity does not originate in the computer model but in the person programming it. This is the case in the coil winder example, where a fruit peeler and a fishing reel are included in the knowledge base among other products. According to the example, using prongs to secure the bobbin can be seen as creativity originating in the experience from the fruit peeler. Innovation is possible in the same way creativity is using analogy design techniques. However, the example given by McAdams et al. does not include any innovative variable settings.
\begin{figure} [h]
    \centering
    \includegraphics [width=0.25\textwidth]{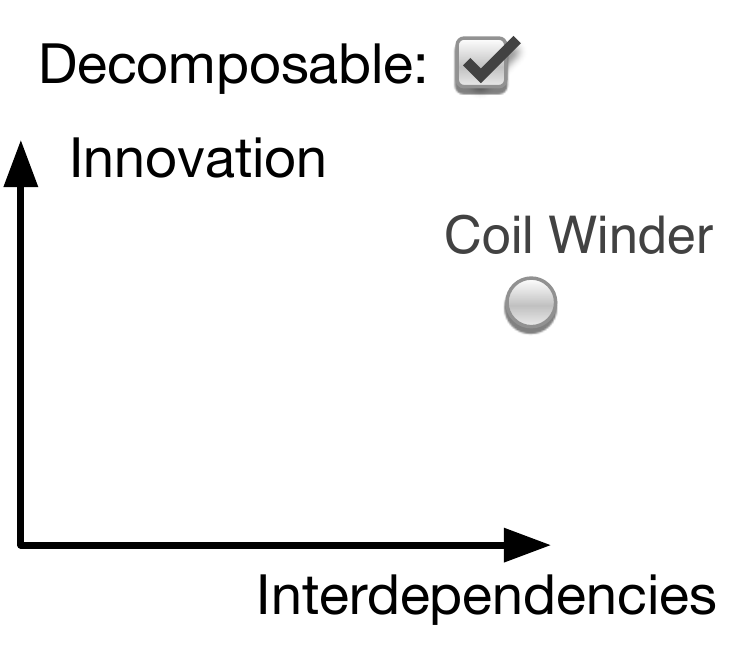}
    \caption{Categorization of the Coil Winder Example}
    \label{cat3.pdf}
\end{figure}

\smallskip
\noindent \emph{Potential AI Approaches:}
In AI, case-based reasoning has always been a successful approach to analogy-based design~\cite{goel:2005,maher:1997}: First, a function is defined which captures the similarity between designs. Then for a new design, a similar old design is identified and adapted to the new requirements. In the last years, generative neural networks offer a new analogy-based solution approach~\cite{doi:10.1080/00207543.2019.1662133,WANG2019105683}: A neural network uses examples to learn the mapping from requirements to product designs and is able to use this learned knowledge to generate new designs for new requirements.

\subsection{Circuit Design}

Feldman et al.~\cite{Feldman.2019} proposed an approach for Boolean circuit design, where circuits are generated using quantified Boolean formulas. Their algorithms solved two proposed problems. While the first problem was generating a Boolean circuit given a circuit topology and the desired requirements, the second problem also included the automatic generation of the circuit topology.

\begin{figure} [h]
    \includegraphics [width=0.5\textwidth]{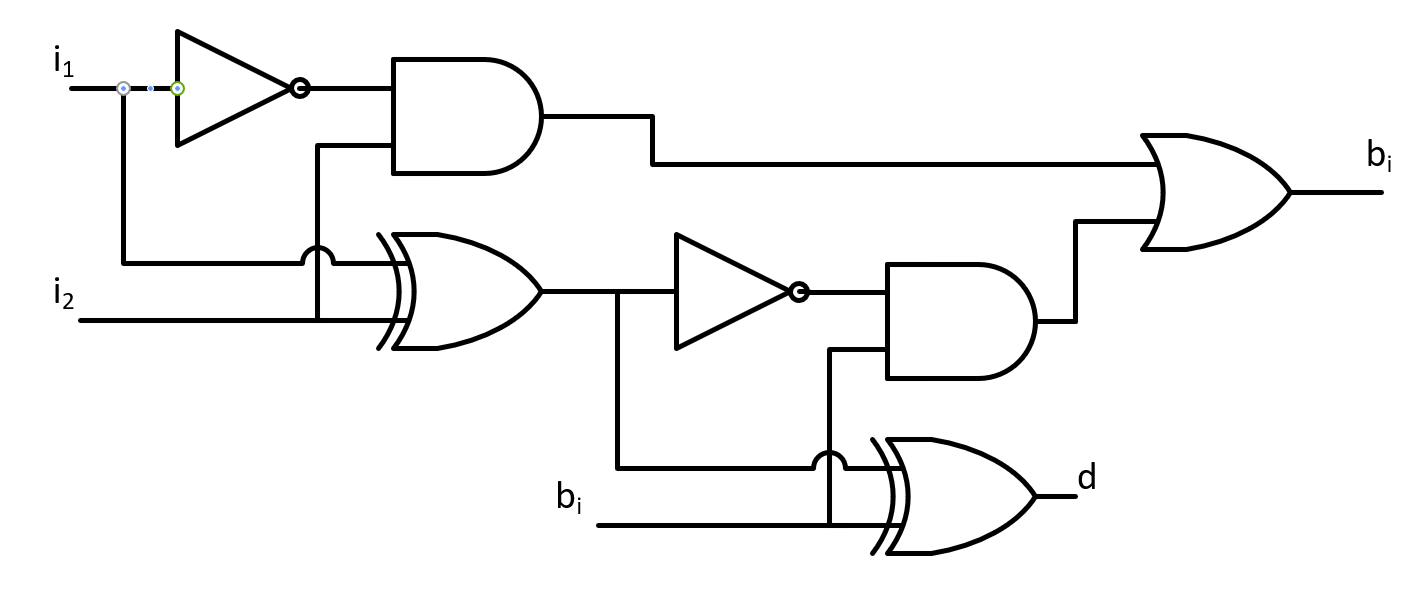}
    \caption{Full substractor based on~\cite{Feldman.2019}}
    \label{fig:fs}
\end{figure}

\smallskip
\noindent \emph{Classification into Categories:} 
The approach falls in the category of functional-based synthesis. Considering the decomposability of boolean circuits, it becomes clear that the topology of these circuits represents a function structure. These structures can be decomposed into subfunctions which are constituted by circuit components like, e.g. gates. Aside from determining the decomposability, the \(PI\) are also computed by looking at the topology. Each gate can be seen as a vertex. Here vertices have either one input and one output, e.g. identity and negation, or two inputs and one output, e.g. AND, OR, XOR gates. Given a topology the \(PI\) can be computed using equation \ref{pc}. A full substractor is shown in figure \ref{fig:fs}. The \(PI\) amounts to \(5/7\) in this example.

The proposed approach can be seen as innovative, when one considers that the algorithms used are able to find solutions which require a different number of components and different compositions. Here the values for the given variables, i.e. number of certain components, may be shifted towards prior unknown values. Which gives the approach innovative capabilities. Since the algorithms are not able to find new variables i.e. the logic gates used to build Boolean circuits, the approach shown by Feldman et al. is not capable of being creative.
\begin{figure} [h]
    \centering
    \includegraphics [width=0.25\textwidth]{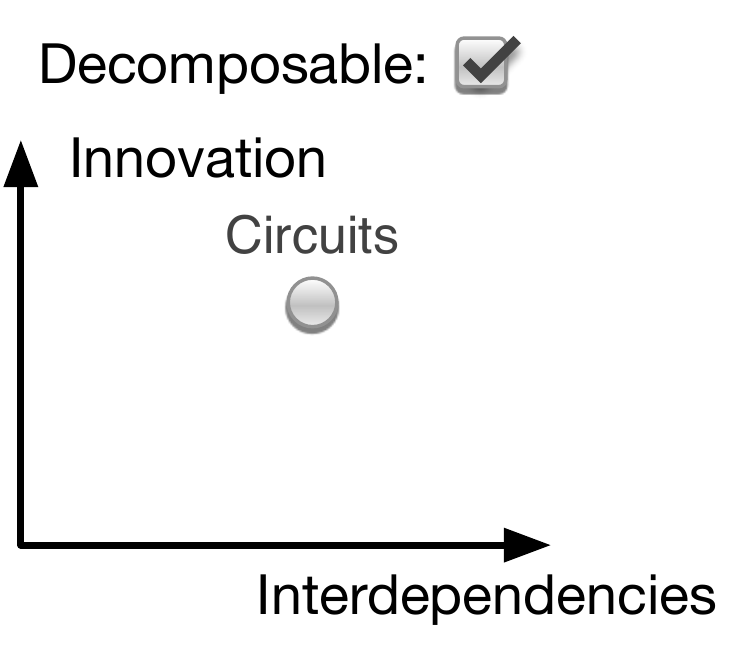}
    \caption{Categorization of the Circuit Example}
    \label{cat4.pdf}
\end{figure}

\smallskip
\noindent \emph{Potential AI Approaches:}  
In AI, such combinatorial problems are often solved using search algorithms such as constraint systems~\cite{BOWEN199373} or genetic search~\cite{https://doi.org/10.1002/aisy.202000075}. Such solutions are helpful if the search space structure is well known.

\section{Conclusion}

Sections \ref{terms} and \ref{class} answer RQ 1. Further, design problems can be categorized by their decomposability. Closely tied to decomposability are the problem interdependencies. Meaningful statements about the \(PI\) can only be made if the problem is decomposable into a function structure. The two exceptions are trivial design problems which can be solved with fundamental concepts and nondecomposable problems, for which no \(PI\) statement can be made apart from viewing the problem as a black box. The remaining two characteristics are concerned with the novelty of nonroutine design problems. While innovation occurs by altering variables to %previously 
unknown values, creativity goes even further by implementing %entirely 
new variables. These characteristics correspond to RQ 2.

\begin{figure} [h]
    \centering
    \includegraphics [width=0.3\textwidth]{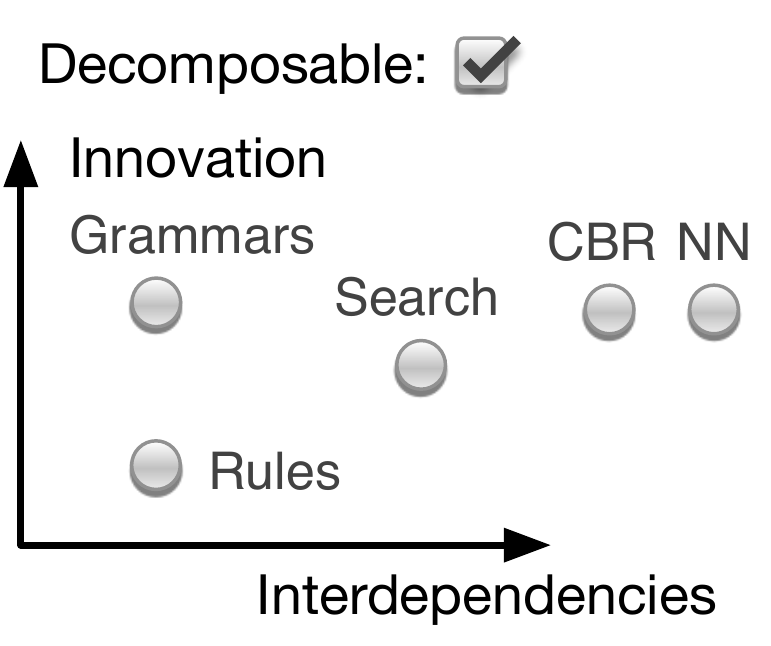}
    \caption{Mapping AI methods to design problem categories.}
    \label{cat5.pdf}
\end{figure}

In relation to RQ 3, the exemplary use cases from the three main directions of AI in product design point out that problems need to have specific characteristics in order to be tackled by these methods. Figure \ref{cat5.pdf} uses the new categories to classify AI technologies for product design. The analysis is based on the examples where each example has been analysed in terms of characteristics and AI solutions.

As long as a problem is decomposable, functional synthesis, design grammars and analogy design approaches are all suited for design problem solving. All approaches are suited to deal with problem interdependencies, given the problem can be mapped to a function structure.

Innovation may be achieved with all mentioned approaches. Altering variable values can be easily done by design grammars by implementing transformation rules which allow bypassing or resetting constraints. In functional-based synthesis innovation is also possible by altering values of variables, which the circuit design example showed. Analogy approaches are less suited for innovation. Here the similarity measurement puts the solutions as close as possible to already existing ones. This leaves less room for assigning unknown values to existing variables. For all approaches there needs to be some kind of measurement for when the alteration led to value. This may be done with a cost function, e.g. less components lead to cheaper production and easier maintenance. Interestingly none of the approaches is able to implement creativity. 

This answers the question asking for problem characteristics which are not handled sufficiently by AI methods. The main reason these methods cannot handle creativity is that all approaches refer to a knowledge base which is externally implemented and assumed to be suited best to the given problems by the programmer. Design grammars are limited to their vocabulary, functional-based approaches rely on a knowledge base and analogy approaches refer to an analogy or case base.

Creativity in product design is what is responsible for disruptive advances, be it the development of machines, software, the internet or even AI itself. Giving the computer the ability to find creative solutions to design problems would be of great assistance for finding solutions for novel design problems. %These could so far only effectively be tackled by experienced engineers or by chance. %Being less dependant on human ingenuity and 
Being able to produce creative solution on a reliable base would greatly improve the product design process and save costs in the development of novel products. Researching creativity could start out on the psychological level to prove new concepts. The ultimate goal should be to achieve historical creativity.

%\printbibliography

\raggedright
\bibliographystyle{abbrv}
\bibliography{BibRosenthal}

\end{document}